\documentclass[11pt]{article}

\usepackage[final]{acl}
\usepackage{times}
\usepackage{latexsym}
\usepackage[T1]{fontenc}
\usepackage[utf8]{inputenc}
\usepackage{microtype}
\usepackage{inconsolata}
\usepackage{graphicx}
\usepackage{amsmath}
\usepackage{graphicx}
\usepackage{booktabs}
\usepackage{amsmath}
\usepackage{amssymb}

\usepackage{listings}
\usepackage{tcolorbox}
\definecolor{deepred}{rgb}{0.631,0.102,0.102}
\definecolor{gyellow}{HTML}{F4B400}
\definecolor{mildyellow}{HTML}{FFF2CC}
\usepackage{colortbl}
\usepackage{float}
\usepackage{placeins}

\title{\includegraphics[scale=0.08]{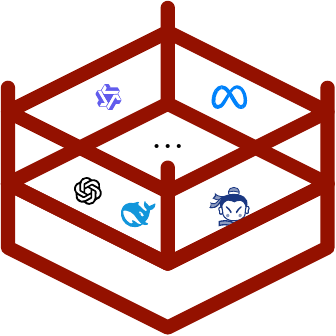} \textsc{GuessArena}: Guess Who I Am? A Self-Adaptive Framework for Evaluating LLMs in Domain-Specific Knowledge and Reasoning}

\author{
  Qingchen Yu\textsuperscript{\rm 1}\thanks{Equal contribution.} \quad
  Zifan Zheng\textsuperscript{\rm 2}\footnotemark[1] \quad
  Ding Chen\textsuperscript{\rm 3}\footnotemark[1] \\
  \textbf{Simin Niu}\textsuperscript{\rm 4} \quad
  \textbf{Bo Tang}\textsuperscript{\rm 1} \quad
  \textbf{Feiyu Xiong}\textsuperscript{\rm 1} \quad
  \textbf{Zhiyu Li}\textsuperscript{\rm 1}\thanks{Corresponding authors: \texttt{lizy@iaar.ac.cn}}\\
  \\
  \textsuperscript{\rm 1} MemTensor (Shanghai) Technology Co., Ltd. \quad
  \textsuperscript{\rm 2} University of Sydney \\
  \textsuperscript{\rm 3} Research Institute of China Telecom \quad
  \textsuperscript{\rm 4} Renmin University of China
}

\begin{document}
\maketitle

\begin{abstract}
The evaluation of large language models (LLMs) has traditionally relied on static benchmarks, a paradigm that poses two major limitations: (1) predefined test sets lack adaptability to diverse application domains, and (2) standardized evaluation protocols often fail to capture fine-grained assessments of domain-specific knowledge and contextual reasoning abilities. To overcome these challenges, we propose \textsc{GuessArena}, an adaptive evaluation framework grounded in adversarial game-based interactions. Inspired by the interactive structure of the \textit{Guess Who I Am?} game, our framework seamlessly integrates dynamic domain knowledge modeling with progressive reasoning assessment to improve evaluation fidelity. Empirical studies across five vertical domains—finance, healthcare, manufacturing, information technology, and education—demonstrate that \textsc{GuessArena} effectively distinguishes LLMs in terms of domain knowledge coverage and reasoning chain completeness. Compared to conventional benchmarks, our method provides substantial advantages in interpretability, scalability, and scenario adaptability. This work provides a scalable and domain-aware solution for LLM evaluation, with the implementation publicly available at \url{https://github.com/IAAR-Shanghai/GuessArena}.
\end{abstract}

\section{Introduction}

\begin{figure}[ht]
    \centering
    \includegraphics[width=0.95\columnwidth]{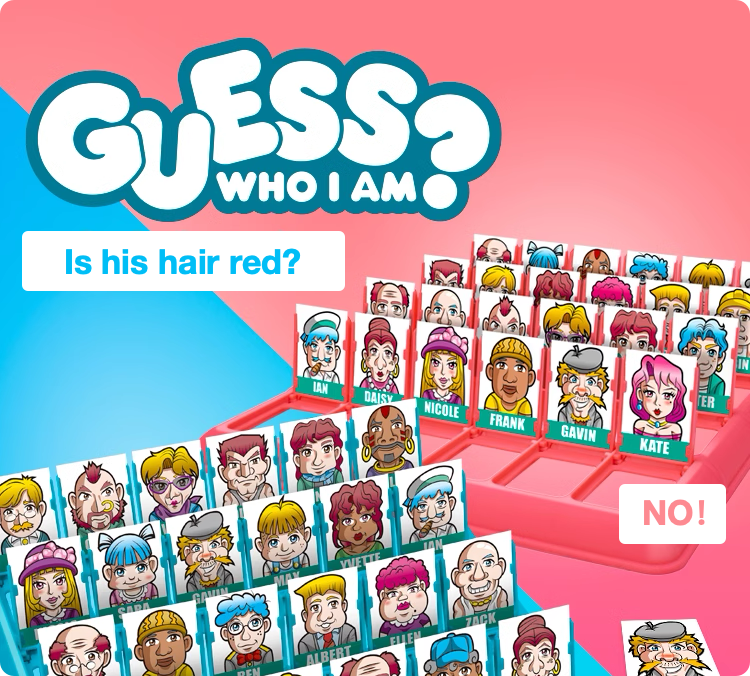}
    \caption{Illustration of the "Guess Who I Am?" game. In this game, two players engage in an interactive process of questioning and reasoning to identify the opponent's chosen card. The player who correctly guesses the target card in the fewest attempts is the winner.}
    \label{fig:gwia_demo}
\end{figure}

The rapid advancement of large language models (LLMs) has driven their widespread adoption across vertical domains such as healthcare, finance, and education~\cite{few-shot_20_NeurIPS_openai, liu2024deepseek, verma2025framework}. However, with the continuous emergence of domain-specific applications—such as financial risk assessment and medical diagnosis—systematically evaluating an LLM’s proficiency in domain knowledge and reasoning ability remains a significant challenge~\cite{chang2024survey,cao2025toward}.

Current mainstream evaluation methods predominantly rely on static benchmark tests (e.g., MMLU~\cite{MMLU_21_ICLR_UCB} and Big-Bench~\cite{big-bench_22_arXiv_Google}), which suffer from two fundamental limitations. First, predefined general-purpose test sets lack the flexibility to dynamically adapt to the specialized assessment requirements of diverse domains. Second, standardized evaluation protocols provide limited fine-grained quantitative analysis of domain-specific contextual reasoning capabilities. 

More critically, when developers seek to construct customized evaluation benchmarks for emerging fields such as blockchain technology and biopharmaceuticals, they often encounter a costly and time-consuming process involving test scenario selection, question annotation, and evaluation pipeline design. This complexity creates a significant barrier to efficient and scalable domain-specific evaluation of LLMs.

Moreover, the limitations of existing evaluation methods extend beyond efficiency concerns. Traditional static benchmarks (e.g., ARC~\cite{ARC_18_arXiv_AI2}) are vulnerable to evaluation biases caused by training data leakage~\cite{LLMCheater_23_arXiv_RUC,xFinder_24_ICLR_IAAR}. In contrast, emerging dynamic evaluation frameworks (e.g., Chatbot Arena~\cite{LMSYS_24_arXiv_UCB}) improve evaluation flexibility through human interaction; however, their results remain inherently influenced by subjective judgments, posing challenges to standardization. Recently, GameArena~\cite{GameArena_24_arXiv_UCSD} proposed a gamified evaluation mechanism, offering a novel approach to assessing general logical reasoning. Nevertheless, its design primarily targets logic-based reasoning tasks and does not adequately address the critical challenge of domain-specific knowledge evaluation.

To address these challenges, we propose \textsc{GuessArena}, an adaptive framework for evaluating domain-specific knowledge and reasoning. As illustrated in Figure~\ref{fig:gwia_demo} and further elaborated in Appendix~\ref{sec:guess_who_i_am}, it transforms the classic "Guess Who I Am" game into a structured evaluation of LLM adaptability in specialized scenarios. The core evaluation pipeline comprises two key stages:

\begin{itemize}
    \item \textbf{Domain Knowledge Modeling}. Automatically processes user-provided domain documents (e.g., medical guidelines, legal statutes, financial reports) to construct a candidate card repository for evaluation.
    \item \textbf{Interactive Reasoning Evaluation}. Simulates real-world decision-making scenarios through a multi-turn dialogue mechanism. By analyzing the model’s questioning strategies and reasoning pathways, the system quantitatively evaluates knowledge retrieval efficiency and logical reasoning effectiveness.
\end{itemize}

\begin{figure*}[t]
    \centering
  \includegraphics[width=0.99\linewidth]{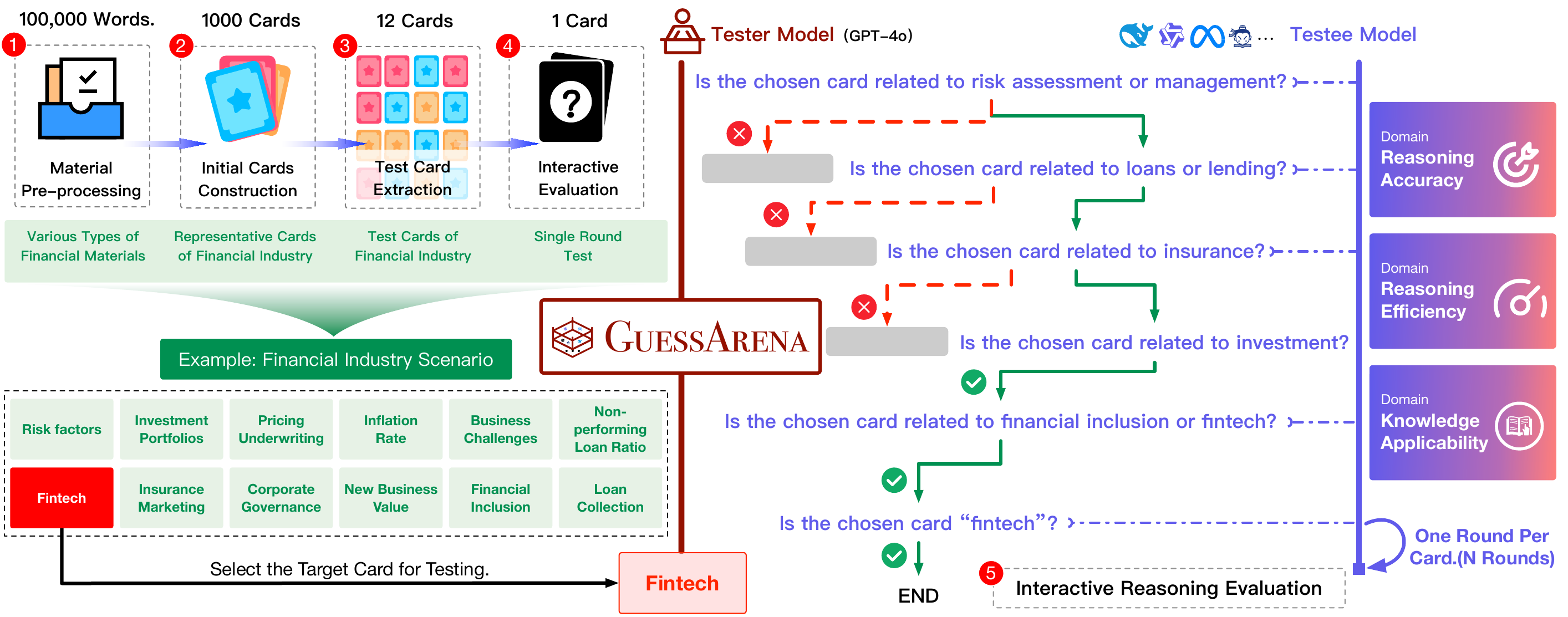} 
    \caption{Framework of \textsc{GuessArena}. The framework comprises two core components: \textbf{Domain Knowledge Modeling} (Left Panel), which parses and models domain-specific documents to generate a candidate card repository for evaluation; and \textbf{Interactive Reasoning Evaluation} (Right Panel), which employs a multi-turn dialogue mechanism to construct an interactive reasoning game, systematically assessing the model's key capability metrics.}
  \label{fig:framework}
\end{figure*}

Compared to existing methods, \textsc{GuessArena} offers the following key contributions:

\begin{itemize}
    \item \textbf{An interactive, reasoning-based, domain-adaptive evaluation framework.} We formalize the mechanics of the "Guess Who I Am" game into a two-stage paradigm—dynamic knowledge modeling and progressive reasoning assessment—seamlessly integrating domain knowledge testing and complex reasoning evaluation within a unified framework.
    
    \item \textbf{An adaptive card extraction algorithm for domain knowledge modeling.} We design an algorithm that automatically extracts structured evaluation cards from unstructured documents (e.g., PDF, HTML, plain text) relevant to the target domain, significantly reducing the cost and effort of building domain-specific evaluation pipelines.
    
    \item \textbf{Comprehensive evaluation across five key industries.} We demonstrate the applicability of \textsc{GuessArena} by evaluating state-of-the-art LLMs in finance, healthcare, manufacturing, information technology, and education. Furthermore, we open-source the entire evaluation framework and benchmark dataset to facilitate future research.
\end{itemize}

\section{Related Work}

\paragraph{Reasoning Evaluation for LLMs}
Existing reasoning evaluation methods primarily rely on carefully designed static benchmark datasets, which often focus on a single type of reasoning task. For example, datasets such as BIG-Bench~\cite{big-bench_22_arXiv_Google}, HotpotQA~\cite{yang2018hotpotqa}, and MMLU~\cite{MMLU_21_ICLR_UCB} are used to evaluate general knowledge reasoning, while MATH~\cite{hendrycks2021measuring} and GSM8K~\cite{GSM8K_21_arXiv_OpenAI} evaluate mathematical and arithmetic reasoning. Similarly, HumanEval~\cite{chen2021codex} and CS-Bench~\cite{song2024cs} are designed to evaluate code reasoning capabilities.

However, these static benchmarks are prone to data contamination and may quickly become obsolete as model capabilities advance~\cite{LLMCheater_23_arXiv_RUC,xFinder_24_ICLR_IAAR}, thus failing to effectively reflect real-world reasoning performance. To address these limitations, researchers have proposed dynamic evaluation approaches~\cite{GameArena_24_arXiv_UCSD,yu2024turtlebench, zhang2024eda}, such as GameArena~\cite{GameArena_24_arXiv_UCSD}, which evaluates LLMs through interactive human-in-the-loop gameplay. While GameArena enables fine-grained evaluation, its reliance on human feedback introduces subjectivity and limits scalability, reducing overall evaluation efficiency. 

In contrast, \textsc{GuessArena} provides a more automated, reproducible, and flexible evaluation framework. By leveraging adaptively generated domain knowledge cards and multi-turn interactive evaluations, it effectively evaluates LLM reasoning capabilities and domain knowledge utilization in specialized and real-world domains.

\paragraph{Domain Knowledge Evaluation}
As LLMs become increasingly integrated into various vertical industries, evaluating their domain-specific knowledge capabilities has become a critical challenge~\cite{chen2024overview,ge2024openagi}. Traditional domain knowledge assessment methods typically rely on manually curated benchmark datasets to measure a model’s proficiency in specific fields~\citep{chang2024survey,yang2024financialknowledge,kim2025medical,liu2023verilogeval}.

For example, Fin-Eva~\cite{fin-eva2025} serves as a financial domain benchmark, covering scenarios such as wealth management, insurance, and investment research. Similarly, MedJourney~\cite{khandekar2024medcalc} evaluates LLM effectiveness in clinical settings, while Shopping MMLU~\cite{jin2024shopping} provides an e-commerce evaluation benchmark based on Amazon shopping data. However, constructing such benchmarks for a new domain is both complex and time-consuming, and static benchmarks face inherent limitations in long-term relevance~\cite{liu2024automatic,boyeau2024autoeval}. In contrast, \textsc{GuessArena} introduces a more generalizable evaluation framework, enabling rapid assessment of LLM performance across different specialized domains without the need for extensive manual dataset construction.

\section{Methodology}

We propose a novel evaluation framework, illustrated in Figure~\ref{fig:framework}. \textsc{GuessArena} supports the construction of a domain knowledge base from user-defined documents, followed by a multi-turn interactive evaluation process to evaluate the knowledge and reasoning abilities of LLMs. To elaborate, Section~\ref{sec:dkbc} introduces the methodology for constructing the domain knowledge base. Then, Section~\ref{sec:iep} details the design of the interactive evaluation process. Finally, Section~\ref{sec:em} outlines the evaluation metrics employed to quantitatively measure model performance within our framework.

\subsection{Domain-oriented Cards Construction} \label{sec:dkbc}

We first extract structured text units from unstructured domain documents and then apply RAG (Retrieval-Augmented Generation) to generate an initial keyword set $\mathcal{K}_0$. The prompt template used in this step is provided in Appendix~\ref{sec:gen_decks_of_cards}.

\begin{equation} Q_d = \text{Template}(d_{\text{meta}}, \mathcal{T}) \end{equation}

Here, $d_{\text{meta}}$ denotes the document metadata, and $\mathcal{T}$ refers to the predefined domain-specific terminology dictionary. For each document, we employ GPT-4o as the retrieval-augmented generator $\mathcal{M}_{\text{RAG}}$ to produce a keyword set by leveraging the document content:

\begin{equation} 
    \mathcal{K}_d = \mathcal{M}_{\text{RAG}}(Q_d \mid d_{\text{content}}) 
\end{equation}

However, the initially extracted keywords may include irrelevant or semantically redundant terms. To refine the keyword set, we further process the candidates by computing their semantic similarity to the evaluation topic. Specifically, we utilize the \texttt{paraphrase-multilingual-MiniLM-L12-v2} model~\cite{Sentence-BERT_19_ACL_UKP} to embed each keyword. Keywords with cosine similarity scores outside a predefined threshold range are filtered out as follows:

\begin{equation}
    f_{\text{filter}}(k_i) = \mathbb{I}\left[\tau_l < \frac{\langle \phi(k_i), \phi(t) \rangle}{\|\phi(k_i)\| \cdot \|\phi(t)\|} < \tau_u\right]
\end{equation}

Here, $\phi(\cdot)$ denotes the Sentence-BERT encoder~\cite{reimers2019sentence}, and the thresholds $\tau_l = 0.35$ and $\tau_u = 0.9$ are empirically determined via grid search. Following this filtering process, we obtain a domain-specific test deck, where each card corresponds to a domain-relevant noun, technical term, or other key concept.

Finally, we apply the spectral clustering algorithm to group the remaining keywords into 10 distinct categories, thereby constructing the domain knowledge base. We begin by computing a similarity matrix $S$, where each entry is defined as $S_{ij} = \cos(\mathbf{v}_i, \mathbf{v}_j)$, with $\mathbf{v}_i$ and $\mathbf{v}_j$ denoting the embedding vectors of keywords $k_i$ and $k_j$, respectively. Next, we derive the normalized Laplacian matrix $L = D^{-1/2} S D^{-1/2}$ and perform spectral decomposition. The resulting eigenvectors are then clustered using the $k$-means algorithm to obtain 10 keyword clusters. During evaluation, users can sample a fixed number of cards from each cluster to construct a test set that ensures comprehensive coverage of domain-specific knowledge while preserving topical diversity.

\subsection{Interactive Evaluation Procedure} \label{sec:iep}

The core of the \textsc{GuessArena} framework is inspired by the classic game "Guess Who I Am?" and incorporates a multi-turn interactive evaluation process. At the beginning of each evaluation session, $N$ cards are sampled from the pre-constructed domain knowledge base to form the evaluation set $\mathcal{D} = \{c_1, c_2, \dots, c_N\}$. In each round $i$, the card $c_i \in \mathcal{D}$ is designated as the target card $g$, which the evaluated LLM must identify. The model undergoes $N$ such rounds, each corresponding to a different target card from the evaluation set.

In each evaluation round, an additional judge model is required alongside the evaluated model. We adopt GPT-4o for this role. The prompt templates for both the judge model and the evaluated model are provided in Appendix~\ref{sec:Prompts_Testers_Testees}. For the judge model, the full evaluation card set $\mathcal{D}$ and the current round’s target card $g$ are given as input. The judge model is constrained to respond strictly with one of four tokens: \texttt{Yes}, \texttt{No}, \texttt{Invalid}, or \texttt{End}, in reply to the evaluated model’s queries and guesses.

The evaluated model is responsible for devising a questioning strategy based on the attributes of the cards in $\mathcal{D}$ and iteratively incorporating feedback from the judge model to infer the target card $g$ using the fewest possible queries. Each evaluation round terminates under one of the following two conditions: (1) the model reaches the maximum number of allowed turns $N$; or (2) the model submits a final guess for the target card $p$.

\subsection{Evaluation Metrics} \label{sec:em}

To comprehensively evaluate the knowledge capability and reasoning ability of the tested model within a specific domain, we design a composite score that integrates the model’s domain reasoning accuracy ($E$), reasoning efficiency ($F$), and knowledge applicability ($K$). The composite score is computed as follows:

\begin{equation}
  \text{score} = w_1 \cdot E + w_2 \cdot F + w_3 \cdot K
  \label{eq:score}
\end{equation}

Here, the weight parameters $w_1$, $w_2$, and $w_3$ control the contribution of each component to the composite score. To ensure that each component contributes equally, we set the weights as $w_1 = w_2 = w_3 = \frac{1}{3}$.

Reasoning accuracy ($E$) is defined as the proportion of correctly guessed target cards in all evaluation rounds. This metric captures the model’s core reasoning correctness and is calculated as:

\begin{equation}
  E = \frac{\text{correct guesses}}{\text{total guesses}}
  \label{eq:effectiveness}
\end{equation}

Here, \textit{correct guesses} refers to the number of times the model correctly identifies the target card, while \textit{total guesses} denotes the total number of guesses made throughout the evaluation. A higher value of $E$ indicates greater accuracy of the model in performing reasoning tasks.

Reasoning efficiency ($F$) quantifies the number of steps the model takes during the reasoning process. It reflects not only the step count but also the model’s capability to quickly narrow down the candidate set and identify the correct card using the fewest possible steps and questions, given all available card information. The reasoning efficiency is computed as follows:

\begin{equation}
  F = \frac{1}{1 + \exp\left(4 \cdot \frac{t_{\text{model}} - t_{\text{rand}}}{t_{\text{rand}}}\right)}
  \label{eq:efficiency}
\end{equation}

Here, $t_{\text{model}}$ denotes the number of reasoning steps taken by the model, $t_{\text{rand}}$ represents the number of reasoning steps required by a random baseline, and $\alpha$ is a hyperparameter controlling the efficiency penalty. A higher value of $F$ indicates that the model completes the task in fewer steps, thereby demonstrating greater reasoning efficiency.

Knowledge applicability ($K$) quantifies the model’s effective utilization of domain knowledge during the reasoning process. This metric penalizes reasoning steps that exceed those of a random baseline, thereby encouraging the model to leverage domain knowledge efficiently within a reasonable number of steps. The knowledge applicability is computed as follows:

\begin{equation}
  K = \exp\left(-\max\left(0, \frac{t_{\text{model}} - t_{\text{rand}}}{t_{\text{rand}}}\right)\right)
  \label{eq:knowledge_utilization}
\end{equation}

The overall score combines reasoning accuracy, reasoning efficiency, and knowledge applicability via a weighted average. A higher score indicates that the model accurately infers the target card while minimizing the number of reasoning steps and effectively leveraging domain knowledge. This comprehensive evaluation method facilitates a more precise assessment of the model’s overall capability in domain-specific reasoning tasks.

\section{Experiments}

\subsection{Experimental Setup}

\paragraph{Domain Datasets}
\textsc{GuessArena} evaluates the performance of LLMs in five specific industries: finance, healthcare, manufacturing, information technology, and education. Specifically, we collected documents from the internet related to these five industries, constructed corresponding domain knowledge bases, and extracted 30 cards from each knowledge base as the evaluation set. The detailed composition of each evaluation set can be found in Appendix~\ref{subsec:exp_setup_details}. 

\paragraph{Evaluated Models}
Based on these evaluation sets, we evaluate nine mainstream top LLMs: GPT-4o~\cite{openai_gpt4o}, OpenAI-o1~\cite{openai_o1}, Claude-3.5-Sonnet~\cite{anthropic_claude}, DeepSeek-V3~\cite{liu2024deepseek}, DeepSeek-R1~\cite{deepseek_r1}, Qwen2.5 (32B-Instruct, 72B-Instruct)~\cite{yang2024qwen2}, Llama-3.3-70B-Instruct~\cite{meta_llama_3_3}, and QwQ-32B~\cite{qwq32b}. Detailed information about each model is shown in Table~\ref{tab:models}.

\begin{table}[htbp!]
\centering
\begin{tabular}{lcccc}
     \toprule
     \textbf{Model} & \textbf{\#Para.} & \textbf{Type} & \textbf{Date} \\ \midrule
     GPT-4o & NaN  & Chat & 2024.05   \\
     OpenAI-o1 & NaN  & Chat  & 2024.09   \\
     Qwen2.5-32B & 32B & Instruct   & 2024.09   \\
     Qwen2.5-72B & 72B  & Instruct  & 2024.09   \\
     Claude-3.5-Sonnet   & NaN  & Chat   & 2024.10   \\
     Llama3.3-70B & 70B & Instruct   & 2024.12   \\
     DeepSeek-V3 & 671B  & Chat   & 2024.12   \\ 
     DeepSeek-R1 & 671B  & Chat   & 2025.01   \\ 
     QwQ-32B & 32B  & Chat   & 2025.03 \\ \bottomrule
\end{tabular}
\caption{LLMs evaluated in this study, ordered by public release date. “\#Para.” listsists the provider-announced parameter count in billions (\textit{B}); “Type” distinguishes models released with \textit{Chat} or \textit{Instruct} interaction styles. \textsc{NaN} indicates that the parameter count has not been publicly disclosed.}
\label{tab:models}
\end{table}

\paragraph{Prompting Strategies}
Three prompting approaches are adopted: \textbf{basic prompt}, \textbf{cot prompt}, and \textbf{knowledge-driven prompt}. The cot prompt guides the model to perform step-by-step reasoning when answering questions, compared to the basic prompt. The knowledge-driven prompt provides the model with background knowledge relevant to the domain evaluation set. Specific prompt templates can be found in Appendix~\ref{sec:Prompts_Testers_Testees}. 

The core hypothesis behind the design of these three prompting strategies is that a model's suboptimal performance in specific domains may stem from either insufficient reasoning ability or a lack of relevant domain knowledge. Therefore, for models with weaker reasoning capabilities, the cot prompt is expected to enhance their reasoning performance. In contrast, for models lacking sufficient domain knowledge, providing background knowledge through knowledge-driven prompts can help improve their overall task performance.

\subsection{Results and Analysis} \label{sec:results}

\begin{table*}[!ht]
\centering
\resizebox{0.9\textwidth}{!}{
\begin{tabular}{@{}lcccccc@{}}
    \toprule
     & \textbf{Info Tech} & \textbf{Finance} & \textbf{Education} & \textbf{Healthcare} & \textbf{Manufacturing} & \textbf{Avg.} \\ \midrule
    Claude-3.5-Sonnet & 0.8535 & 0.7941 & 0.8487 & 0.9134 & 0.8442 & 0.8508 \\
    DeepSeek-R1 & 0.8739 & 0.7855 & 0.8314 & 0.8106 & 0.8025 & 0.8208 \\
    DeepSeek-V3 & 0.8988 & 0.8016 & 0.8749 & 0.9279 & 0.7974 & 0.8601 \\
    GPT-4o & \textbf{0.9244} & 0.8465 & \underline{0.9020} & \textbf{0.9302} & \textbf{0.9043} & \underline{0.9015} \\
    Llama-3.3-70B-Instruct & 0.8045 & 0.7581 & 0.8047 & 0.7775 & 0.7966 & 0.7883 \\
    OpenAI-o1 & 0.8814 & \textbf{0.9199} & \textbf{0.9271} & \underline{0.9282} & 0.8705 & \textbf{0.9054} \\
    Qwen2.5-32B-Instruct & 0.8808 & \underline{0.8610} & 0.8366 & 0.8323 & 0.8360 & 0.8493 \\
    Qwen2.5-72B-Instruct & \underline{0.9052} & 0.8533 & 0.8933 & 0.9106 & \underline{0.9020} & 0.8929 \\
    QwQ-32B & 0.8543 & 0.8597 & 0.8596 & 0.8902 & 0.8991 & 0.8726 \\ \bottomrule
\end{tabular}
}
\caption{Domain-wise \textsc{GuessArena} scores (higher is better) under the \emph{basic prompt} setting. The table reports composite \textsc{GuessArena} results for nine LLMs across five industry domains; the rightmost column gives the macro-average across domains. The highest score in each column is \textbf{boldfaced}, and the
second-highest is \underline{underlined}.}
\label{tab:basic_prompt}
\end{table*}

\begin{table*}[!ht]
\centering
\resizebox{0.9\textwidth}{!}{
\begin{tabular}{@{}lcccccc@{}}
\toprule
     & \textbf{Info Tech} & \textbf{Finance} & \textbf{Education} & \textbf{Healthcare} & \textbf{Manufacturing} & \textbf{Avg.} \\ \midrule
    Claude-3.5-Sonnet & 0.8956 & 0.8089 & 0.8545 & 0.9097 & 0.8469 & 0.8631 \\
    DeepSeek-R1 & 0.8657 & 0.8333 & 0.8382 & 0.8331 & 0.8292 & 0.8399 \\
    DeepSeek-V3 & 0.8676 & 0.8092 & 0.8332 & 0.8830 & 0.8066 & 0.8399 \\
    GPT-4o & \textbf{0.9149} & 0.8520 & \textbf{0.8974} & \textbf{0.9409} & 0.8748 & 0.8960 \\
    Llama-3.3-70B-Instruct & 0.8263 & 0.7482 & 0.7878 & 0.8231 & 0.8037 & 0.7978 \\
    OpenAI-o1 & 0.8762 & \textbf{0.8932} & 0.8881 & \underline{0.9377} & 0.8885 & \underline{0.8967} \\
    Qwen2.5-32B-Instruct & 0.8722 & 0.8047 & 0.8422 & 0.8612 & 0.8582 & 0.8477 \\
    Qwen2.5-72B-Instruct & \underline{0.9124} & \underline{0.8736} & \underline{0.8943} & 0.9047 & \textbf{0.9033} & \textbf{0.8977} \\
    QwQ-32B & 0.8616 & 0.8386 & 0.8408 & 0.8871 & \underline{0.8918} & 0.8640 \\ \bottomrule
\end{tabular}
}
\caption{Domain-wise \textsc{GuessArena} scores (higher is better) under the
\emph{cot prompt} setting. Composite \textsc{GuessArena} results are shown for nine LLMs across five industry domains; the rightmost column reports the
macro-average across domains. The highest value in each column is \textbf{boldfaced}, and the second-highest is \underline{underlined}.}
\label{tab:cot_prompt}
\end{table*}

As shown in Table~\ref{tab:basic_prompt}, under the basic prompt setting, OpenAI-o1 demonstrates the best overall performance, outperforming all other non-reasoning models evaluated. Qwen2.5-72B-Instruct shows relatively strong performance in the information technology and manufacturing industries. In contrast, Llama-3.3-70B-Instruct performs comparatively worse overall, with particularly low scores in the finance and healthcare industries.

To further verify the effectiveness of the \textsc{GuessArena} method in distinguishing the reasoning and domain knowledge capabilities of different LLMs in specific fields, we designed two prompting strategies: the cot prompt and the knowledge-driven prompt. In the cot prompt strategy, the tested models are guided to perform step-by-step reasoning in order to enhance their reasoning abilities. As shown in Table~\ref{tab:cot_prompt}, the experimental results indicate that for models with weaker reasoning abilities, using the cot prompt strategy leads to significant improvements compared to their performance under the basic prompt setting. Notably, Llama-3.3-70B-Instruct and Claude-3.5-Sonnet achieved higher scores across multiple domains. For example, in the healthcare domain, the performance of Llama-3.3-70B-Instruct improved by 5.86\%, and in the information technology domain, Claude-3.5-Sonnet's score increased by 4.93\%. These results demonstrate that \textsc{GuessArena} can effectively distinguish differences in reasoning abilities of LLMs across vertical domains.

\begin{table*}[!ht]
\centering
\resizebox{0.9\textwidth}{!}{
\begin{tabular}{@{}lcccccc@{}}
\toprule
     & \textbf{Info Tech} & \textbf{Finance} & \textbf{Education} & \textbf{Healthcare} & \textbf{Manufacturing} & \textbf{Avg.} \\ \midrule
    Claude-3.5-Sonnet & \underline{0.8872} & 0.8450 & 0.8703 & 0.8741 & 0.8734 & 0.8700 \\
    DeepSeek-R1 & 0.7822 & 0.8269 & 0.8343 & 0.7892 & 0.7685 & 0.8002 \\
    DeepSeek-V3 & 0.8497 & 0.8065 & \underline{0.9021} & \textbf{0.9256} & 0.8618 & 0.8691 \\
    GPT-4o & \textbf{0.9124} & \underline{0.8835} & \textbf{0.9244} & 0.8997 & 0.8721 & \textbf{0.8984} \\
    Llama-3.3-70B-Instruct & 0.8309 & 0.7652 & 0.7865 & 0.8382 & 0.7872 & 0.8016 \\
    OpenAI-o1 & 0.8736 & \textbf{0.9051} & 0.8899 & \underline{0.9218} & \underline{0.8849} & \underline{0.8951} \\
    Qwen2.5-32B-Instruct & 0.8536 & 0.8011 & 0.8572 & 0.9015 & 0.8580 & 0.8543 \\
    Qwen2.5-72B-Instruct & 0.8856 & 0.8518 & 0.8792 & 0.9133 & \textbf{0.8925} & 0.8845 \\
    QwQ-32B & 0.8612 & 0.8256 & 0.8473 & 0.9064 & 0.8613 & 0.8604 \\ \bottomrule
\end{tabular}
}
\caption{Domain‐wise \textsc{GuessArena} scores (higher is better) under the
\emph{knowledge-driven prompt} setting. Nine LLMs are evaluated with prompts that explicitly inject retrieved domain knowledge across five industry domains; the rightmost column
shows the macro-average over domains. The highest value in each column is \textbf{boldfaced}, and the second-highest is \underline{underlined}.}

\label{tab:know_prompt}
\end{table*}

\begin{figure*}[!ht]
  \centering
  \includegraphics[width=2.0\columnwidth]{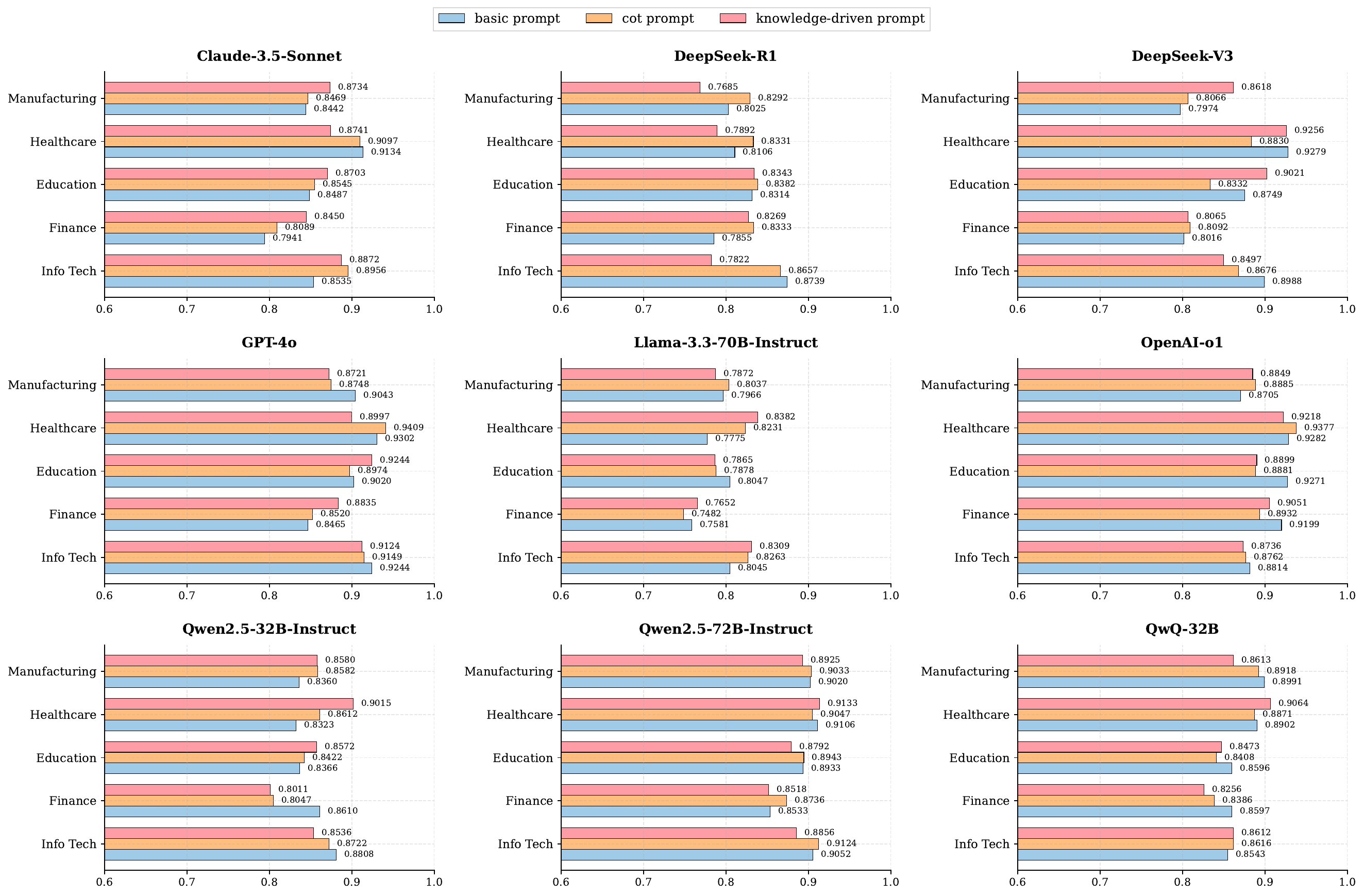}
    \caption{Cross-domain \textsc{GuessArena} scores (higher is better) for nine
    LLMs under three prompting strategies. Grouped bars show the composite \textsc{GuessArena} performance achieved with \emph{basic}, \emph{cot}, and \emph{knowledge-driven} prompts in each of the five industry domains, allowing a visual comparison of prompt effectiveness across models and domains.}
  \label{fig:bar_chart}
\end{figure*}

Under the knowledge-driven prompting strategy, we provide the tested models with customized background knowledge for each domain-specific evaluation set (generated by GPT-4o; the prompt template is shown in Figure~\ref{fig:prompt_gen_kbg} in Appendix~\ref{sec:Prompts_Testers_Testees}). As shown in Table~\ref{tab:know_prompt}, the experimental results indicate that for models lacking corresponding domain knowledge, the provision of relevant background information significantly improves performance, particularly for those that underperformed under the basic prompt setting. For instance, the average score of Claude-3.5-Sonnet across the five vertical domains increased by 2.26\% compared to the basic prompt, while Llama-3.3-70B-Instruct improved by 1.69\%. Notably, both Claude-3.5-Sonnet and Llama-3.3-70B-Instruct showed substantial gains in the information technology, finance, and education domains. These findings demonstrate that equipping models with vertical domain knowledge leads to marked performance improvements, which are effectively captured by the \textsc{GuessArena} scores.

Figure~\ref{fig:bar_chart} presents a comparative bar chart showing the scores of nine LLMs across five domains under three prompting strategies. As illustrated, for stronger models like OpenAI-o1 and GPT-4o, performance differences across prompting strategies are minimal, likely due to their strong reasoning and domain knowledge. In contrast, for other models, if a model has weaker reasoning ability but possesses solid domain knowledge, the cot prompt leads to more significant performance gains; conversely, if a model has strong reasoning ability but lacks sufficient domain-specific knowledge, the knowledge-driven prompt results in notable improvements. For example, in the finance domain, both Claude-3.5-Sonnet and Llama-3.3-70B-Instruct perform significantly better under the knowledge-driven prompt compared to the basic and cot prompts. These experimental results demonstrate that \textsc{GuessArena} effectively distinguishes the reasoning and domain knowledge capabilities of LLMs in vertical domains.

\begin{table*}[!ht]
\centering
\resizebox{0.9\textwidth}{!}{
\begin{tabular}{lcc}
    \toprule
     & \textbf{Agreement with Human Annotations (\%)} & \textbf{Agreement with GPT-4o (\%)} \\
    \midrule
    \textbf{GPT-4o} & \textbf{92.33} & -- \\
    Qwen2.5-72B-Instruct & 88.33 & 88.17 \\
    DeepSeek-V3 & 88.25 & 87.25 \\
    Llama-3.3-70B-Instruct & 88.00 & 85.50 \\
    Claude-3.5-Sonnet & 86.42 & 85.25 \\
    Qwen2.5-32B-Instruct & 86.08 & 86.75 \\
    \rowcolor{gray!10} Majority Voting & 90.58 & 88.08 \\
    \bottomrule
\end{tabular}
}
\caption{Consistency of different judge models with human annotations and GPT-4o judgments. GPT-4o serves as the primary reference model for judgment. The last row reports results from majority voting over all models.}
\label{tab:judge-model-consistency}
\end{table*}

\begin{figure*}[!ht]
    \centering
    \includegraphics[width=0.9\linewidth]{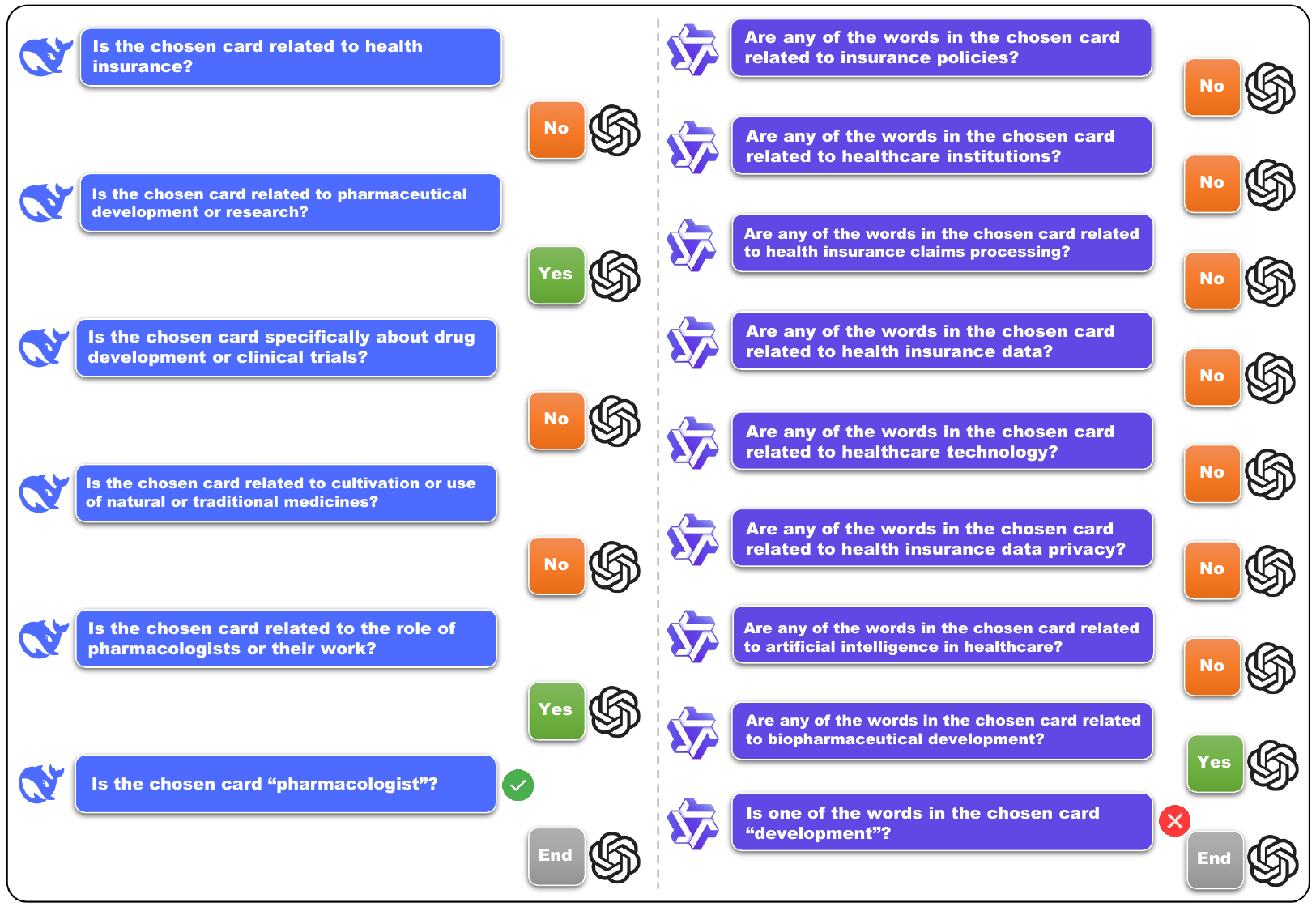}
    \caption{Interactive guessing trajectories in the healthcare scenario. \textbf{DeepSeek-V3} (left) and \textbf{Qwen-2.5-32B-Instruct} (right) pose sequential yes/no questions to identify the target card, \emph{Pharmacologist}. Rounded boxes contain model-generated queries; the colored chips denote the oracle’s feedback (green: Yes, red: No, grey: End).}
    \label{fig:case_study}  
\end{figure*}

\subsection{Further Discussion}

In the experiments described in Section~\ref{sec:results}, we used GPT-4o as the default judge model. To further validate the reliability of GPT-4o in this role, we randomly sampled 1,200 instances from the full evaluation set and invited human annotators to provide gold-standard labels. We then designated several mainstream LLMs, including GPT-4o, as judge models and measured their agreement with the human annotations. 

The results, summarized in Table~\ref{tab:judge-model-consistency}, show that GPT-4o attained the highest agreement rate with human annotations at 92.33\%, underscoring its stability and reliability in judgment tasks. Other judge models also performed comparably, all achieving agreement rates above 86\%. Notably, Qwen2.5-72B-Instruct (88.33\%) and DeepSeek-V3 (88.25\%) demonstrated strong alignment with human judgments. We also calculated a “majority voting” result based on the predictions of all models excluding GPT-4o. This method achieved an agreement of 90.58\% with the human annotations and 88.08\% with GPT-4o’s judgments.

In summary, GPT-4o exhibits strong reliability as a judge model in~\textsc{GuessArena}, and other leading LLMs also produce agreement levels comparable to human annotations. These results suggest that the choice of judge model has a relatively limited impact on the final evaluation outcomes, thereby supporting the credibility and generalizability of our experimental conclusions.

\subsection{Case Study}

From the experimental results above, it can be observed that the DeepSeek-V3 model overall demonstrates good reasoning ability and logical thinking, while Qwen2.5-32B-Instruct performs relatively poorly in the~\textsc{GuessArena} evaluation, possibly due to limitations in its parameter size. To further clarify the differences between the two models in the evaluation, we selected a case for analysis.

As shown in Figure~\ref{fig:case_study}, for the term "pharmacologist" in the healthcare domain, the DeepSeek-V3 model on the left was able to guess the correct answer after 6 rounds of questions, while the Qwen2.5-32B-Instruct model on the right had a total of 9 rounds of conversation but did not arrive at the correct answer. Referring to the 30 card terms from the healthcare domain displayed in Appendix~\ref{subsec:exp_setup_details}, DeepSeek-V3 first grouped these 30 terms into broad categories. For example, it combined several terms related to health insurance and asked the first question, "Is the chosen card related to health insurance?" After receiving a "No" response, it could eliminate multiple cards from the deck at once. This process continued until it correctly guessed the target card.

On the other hand, Qwen2.5-32B-Instruct appears somewhat clumsy, as it does not efficiently group the 30 terms. After each "No" response, it can only eliminate 1-2 candidate terms and often makes a guess without being confident. Both in practical applications and in LM Arena~\footnote{\href{https://lmarena.ai}{https://lmarena.ai}}, it is evident that DeepSeek-V3 demonstrates much stronger overall reasoning capabilities compared to Qwen2.5-32B-Instruct, and our experimental case aligns with this observation. This indicates that~\textsc{GuessArena} can accurately and effectively test the comparative capabilities of different LLMs.

\section{Conclusion}

The proposed \textsc{GuessArena} framework offers an innovative solution for evaluating LLMs' domain-specific knowledge and reasoning capabilities. By integrating dynamic knowledge modeling with progressive reasoning evaluation, \textsc{GuessArena} adapts to diverse domain evaluation needs and evaluates model performance through multi-turn interactions in realistic scenarios. Compared to traditional static benchmarks, our framework enables more efficient and cost-effective evaluation of domain-specific reasoning capabilities while alleviating credibility concerns arising from question leakage in static benchmarks.

In experiments conducted across five predefined vertical domains, \textsc{GuessArena} effectively revealed performance disparities among state-of-the-art LLMs, particularly in reasoning capability and domain knowledge utilization. By tailoring the evaluation pipeline and strategies, our framework enables fine-grained differentiation of models' reasoning and knowledge competencies within specific domains. Experimental results show that \textsc{GuessArena} not only delivers more detailed insights than traditional benchmarks but also flexibly adapts to diverse domain requirements. Overall, \textsc{GuessArena} provides a reliable, scalable, and highly adaptable framework for domain-specific LLM evaluation, offering a robust foundation for future research and development.

\section*{Limitations}

While our framework demonstrates strong applicability by enabling efficient and low-cost customization of evaluation pipelines for assessing domain-specific reasoning and knowledge capabilities, it may not be suitable for all evaluation scenarios. For instance, tasks such as medical diagnosis or legal argumentation often require open-ended and interpretative reasoning, which may not be fully captured by the current evaluation mechanism.

In addition, our experiments adopt GPT-4o as the default judge model. Although we verify the consistency between its evaluations and those of other LLMs as well as human assessments, potential biases may still arise in long-tail domains due to the judge model's limited domain coverage or inherent preference. Future work could incorporate user-defined judge models and ensemble-based voting strategies to enhance the precision and robustness of evaluations.

Finally, although our framework effectively evaluates and differentiates multiple state-of-the-art LLMs across five predefined vertical domains, further investigations across a broader set of long-tail domains and additional models would provide the community with more comprehensive benchmarks.

\bibliography{ref}

\newpage
\clearpage

\onecolumn
\appendix
\definecolor{mydeepblue}{HTML}{000066}

\section{Fundamentals of the "Guess Who I Am?" Game} \label{sec:guess_who_i_am}

This appendix provides a detailed exposition of the classic two-player deduction board game "Guess Who I Am?". This game serves as the foundational inspiration for our proposed \textsc{GuessArena} framework, with its interactive reasoning paradigm providing a robust basis for evaluating the domain-specific knowledge and reasoning capabilities of LLMs.

\paragraph{Game Components and Setup}
The game typically involves two identical game boards, each featuring a grid of character portraits. These portraits possess distinct and discernible features such as hair color, presence of glasses, or headwear. As illustrated in Figure~\ref{fig:gwia_demo} of the main paper, players sit opposite each other, each with their game board. At the commencement of the game, both players secretly select one character card from an identical deck, placing it in a concealed holder so that it remains unknown to the opponent.

\paragraph{Gameplay Dynamics}
The core of "Guess Who I Am?" lies in its strict turn-based questioning protocol. Players alternate turns posing a single question about a feature of the opponent's secret character. Crucially, these questions must be structured to elicit a definitive "Yes" or "No" response (e.g., "Does your character have red hair?"). Based on the opponent's truthful answer, the querying player employs a process of logical elimination: all character portraits on their own board that do not conform to the newly revealed information are flipped down or marked as irrelevant. This iterative elimination progressively narrows the set of plausible candidate characters.

\paragraph{Strategic Principles and Objective}
The game's primary objective is to identify the opponent's secret character in the fewest possible turns. This implicitly incentivizes players to formulate questions that maximize the reduction of the candidate set in each round, thereby enhancing the efficiency of their deductive reasoning. Once a player is confident in identifying the opponent's secret character, they declare a final guess. A correct guess results in immediate victory for that player, with overall performance typically evaluated by the minimal number of turns or questions required to achieve the correct identification.

\section{Knowledge Base and Card Generation} \label{sec:Construct_GuessArena}

This section primarily details the process of constructing the domain knowledge base and the specific methods for generating card decks based on the given topics using the knowledge base.

\subsection{Sources of Domain Knowledge}

Table~\ref{tab:document_numbers} lists the number of documents we collected for the five industries. These documents were sourced from annual reports, news updates, and other publications of Fortune Global 500 companies.

\begin{table}[htbp!]
\footnotesize
\centering
\begin{tabular}{lccccc}
     \toprule
     \textbf{Category} & \textbf{Education} & \textbf{Finance} & \textbf{Healthcare} & \textbf{Info Tech} & \textbf{Manufacturing} \\ \midrule
     Doc. Number & 25 & 20 & 26 & 29 & 28 \\ 
     \toprule
\end{tabular}
\caption{Number of source documents collected for each industry domain
before card extraction in \textsc{GuessArena}. A total of 128 documents spanning Education, Finance, Healthcare, Information Technology, and Manufacturing are used to build the domain-specific card pools; counts by domain are reported in the table.}
\label{tab:document_numbers}
\end{table}

\subsection{Generating Decks of Cards} \label{sec:gen_decks_of_cards}

Figure~\ref{fig:prompt_gen_kws_deck} illustrates the prompt used for generating card keywords from domain documents. In this figure, the \textcolor{red}{red part} represents the system prompt, while the \textcolor{blue}{blue part} represents the instruction. The same color-coding conventions are applied throughout the appendix.

\begin{figure}[htbp]
\small
\centering
\begin{tcolorbox}
    \textcolor{red}{
    You are an expert in the field of \{name\}, which focuses on \{description\}, and your task is to generate a set of domain-specific keywords based on the provided documents. This keyword set should contain \{max\_keywords\} keywords, each being a technical term or jargon from the field, covering as many knowledge points as possible. Avoid repetition, ensuring diversity and uniqueness among the keywords.
    }
    \\
    \textcolor{blue}{
    When generating the keywords, please follow these guidelines:\\
    1. \textbf{Domain Focus}: Prioritize professional terms, concepts, or industry jargon from the field, ensuring the keywords accurately reflect the core content of the domain.\\
    2. \textbf{Uniqueness}: Ensure that each keyword in the set is unique, avoiding synonyms or near-synonyms. Each keyword must be a complete term, avoiding any abbreviations or combinations with explanatory content. For example, avoid forms like “product lifecycle management (plm)”, “lms (learning management system)”, or similar formats. Keywords should be the full term without any parentheses, such as “product lifecycle management” or “learning management system”.\\
    3. \textbf{Broad Coverage}: The keywords should broadly cover various knowledge aspects of the field, including common terms, basic concepts, and specialized vocabulary from subfields. If appropriate, you may also include well-known entities within the domain, such as company names, product names, and people's names, as these entities are important representatives of the field.\\
    4. \textbf{Contextual Relevance}: Ensure that there is an inherent connection between the keywords. For example, in the finance industry, "capital markets" might be related to "stocks," or "balance sheets" might be connected to "financial analysis." Make sure related keywords have sufficient contextual relevance.\\
    5. \textbf{Diversity and Representativeness}: Ensure that the keywords span different domains, levels, and knowledge points, covering not only basic concepts but also niche terms and specialized language unique to the industry.\\
    \\
    The final keywords should all be in English and separated by semicolons. Please do not include any additional explanations, formatting, or unnecessary text, and return only the list of keywords.\\
    \#\#\#\#Example Output:
    keyword1; keyword2; keyword3; ...; keywordn}
\end{tcolorbox}
\caption{Prompt template for deriving domain-specific keywords that seed the \textsc{GuessArena} card deck.}
\label{fig:prompt_gen_kws_deck}
\end{figure}

\FloatBarrier
\section{Experimental Setup and Prompt Templates} \label{sec:exp_details}

This section provides supplementary details regarding the experiments that were not mentioned in the main text, such as the experimental setup and the prompts used in the experiments.

\subsection{Experimental Setup Details} \label{subsec:exp_setup_details}

In this study, we extracted a total of 5 industries, with 30 keywords for each industry, resulting in 150 keywords for constructing the card deck. The specific keywords for each industry are as follows:

\textbf{Education}: \textit{policy encouragement, policy development, policy implementation, vocational training, education policies, online learning engagement, internet of things in education, social-emotional learning, personalized training services, educational system expansion, online learning communities, employment skills enhancement, online learning platforms, numeracy education, e-learning platforms, educational data mining, knowledge economy, training institution review, textbook publishing, higher education providers, education accessibility, educational success, school safety, homeschooling, early childhood education, social learning, digital learning models, digital learning trends, online learning scalability, self-directed learning.}

\textbf{Healthcare:} \textit{internet of things in healthcare, internet healthcare, healthcare technology, specialty drugs, healthcare data, cell therapy, urban health, newborn care leave, epidemiological method, personal care products, herbal medicine cultivation, biopharmaceutical development, clinical trial support, health insurance data security, pharmacologist, national health insurance, health infrastructure, healthcare infrastructure, health insurance data management, health insurance claims processing, health insurance data privacy, artificial intelligence in healthcare, occupational health, over-the-counter medication, national health commission, health history, healthcare institutions, health insurance policy, health financing, traditional medicine.}

\textbf{Finance:} \textit{risk factors, risk assessment, risk classification, price-to-earnings ratio, non-performing loan ratio, financial investment, corporate governance, pricing underwriting, new business value, financial inclusion, financial product operation, investment portfolios, internal rate of return, loan monitoring, business quality, inflation rate, agricultural finance, equity securities, market inflection points, loan-to-value ratio, fintech, government bond investment, investment risk, loan collection, insurance industry, insurance marketing, business challenges, financial intermediation, alternative investments, retail loan proportion.}

\textbf{Information Technology:} \textit{iot industry, iot market development, iot market, it budgeting, sensor integration, server applications, uxsinodb, information security, ai server vendors, iot architecture, iot platform, nvdia vgpu, ai server integration, data center operations, data security management, communication latency, it infrastructure, it project management, cybersecurity framework, ai servers, pc industry trends, vsmp, offline data collection, data analytics, distributed storage system, task management, data center knowledge, wearable technology, npus, cloud service providers.}

\textbf{Manufacturing:} \textit{material inspection, smt inspection, fpc inspection, green manufacturing, carbon fiber composite production, total productive maintenance, manufacturing investments, composite material supply chain, delivery capability, manufacturing strategy, equipment understanding, manufacturing strategies, textile manufacturing, manufacturing partnerships, automotive equipment manufacturing, manufacturing excellence, volkswagen supplier, energy production, product lifecycle, industrial symbiosis, invention patents, laser cutting, advanced electronic materials, manufacturing limitation, automotive manufacturing, full-process delivery, industrial internet, advanced robotics, business process management, supply chain finance.}

\subsection{Prompt Templates} \label{sec:Prompts_Testers_Testees}

The following describes the prompt templates used in the interactive evaluation of the GuessArena Framework. Figure~\ref{fig:prompt_tester} illustrates the prompt for the judge model (e.g., GPT-4o), which evaluates and responds to queries from the evaluated model. Figures~\ref{fig:prompt_testee_basic}, \ref{fig:prompt_testee_cot}, and \ref{fig:prompt_testee_kbg} depict three types of prompts for the evaluated model to determine its next question.

Among these, the prompt shown in Figure~\ref{fig:prompt_testee_kbg} incorporates domain-specific background knowledge. This background knowledge text is generated using the process illustrated in Figure~\ref{fig:prompt_gen_kbg}.

\begin{figure}[htbp]
\small
\centering
\begin{tcolorbox}
    \textcolor{red}{
    You will participate as a judge in a card-guessing game. The game card deck is: \{deck\_of\_cards\}, and the card secretly chosen in this game is: \{chosen\_card\}. Your duty is to objectively and fairly answer players' questions based on the characteristics and information of the selected card.}
    \\
    \textcolor{blue}{\\
    \#\# Judging Rules\\
    1. \textbf{Response Rules:} You can only use four standard responses: "[Yes]", "[No]", "[Invalid]", and "[End]". Do not provide any additional explanations. Maintain absolute objectivity in your responses and do not reveal any specific information about the chosen card.\\
    2. \textbf{Question Determination:} If a player's question is outside the scope of the card deck or obviously unreasonable, respond with "[Invalid]"; if the question is clear and within the answerable range, respond with "[Yes]" or "[No]" based on the actual characteristics of the chosen card.\\
    3. \textbf{Consistency Principle:} For substantially the same or highly similar questions, you must maintain consistent responses. Ensure that each question receives a fair and accurate judgment.\\
    4. \textbf{Game Termination:} When the player gives the final guess of the chosen card, you need to immediately respond with "[End]".\\
    \\
    \#\# Judge's Responsibilities\\
    - Maintain absolute neutrality, do not provide any additional hints\\
    - Strictly follow the above rules to ensure the fairness of the game process}\\
    \\
    \textcolor{mydeepblue}{\{\textbf{\textit{dialog history}}\}}
\end{tcolorbox}
\caption{Prompt template issued to the \emph{judge model} (oracle judge) in \textsc{GuessArena}.}
\label{fig:prompt_tester}
\end{figure}

\begin{figure}[htbp]
\small
\centering
\begin{tcolorbox}
    \textcolor{red}{
    You will participate as a player in a card-guessing game. In front of you is a deck of cards: \{deck\_of\_cards\}, and your task is to guess the chosen card that has been secretly selected through a series of questions. Your questions can only be answered with "[Yes]" or "[No]".}
    \\
    \textcolor{blue}{\\
    \#\# Game Rules\\
    1. \textbf{Question Rules:} You may ask only one clear and concise question at a time, which must be answerable with "[Yes]" or "[No]". The question must not contain line breaks, nor can it directly ask about the card's specific identity. Do not request additional hints.\\
    2. \textbf{Scoring Mechanism:} The game score is inversely proportional to the number of questions asked. The fewer the questions, the higher the final score, assuming you correctly guess the chosen card. Successfully and quickly identifying the target card is key to achieving a high score.\\
    3. \textbf{Guessing Process:} After each question, wait for the judge's response, then use that information, along with previous questions, to ask the next question. When you make your final guess for the chosen card, the judge will immediately respond with "[End]", regardless of whether the guess is correct.\\
    4. *\textbf{Invalid Behaviors:} Repeating the same question or guess is prohibited. Any questions or guesses unrelated to the game will be marked as "[Invalid]" by the judge.}\\
    \\Please begin by asking your first question.\\
    \\
    \textcolor{mydeepblue}{\{\textbf{\textit{dialog history}}\}}
\end{tcolorbox}
\caption{Prompt template for the \emph{evaluated model} (player) under the \emph{basic} prompting regime in \textsc{GuessArena}.}
\label{fig:prompt_testee_basic}
\end{figure}

\begin{figure}[htbp]
\small
\centering
\begin{tcolorbox}
    \textcolor{red}{
    You will participate as a player in a card-guessing game. In front of you is a deck of cards: \{deck\_of\_cards\}, and your task is to guess the chosen card that has been secretly selected through a series of questions. Your questions can only be answered with "[Yes]" or "[No]".}
    \\
    \textcolor{blue}{\\
    \#\# Game Rules\\
    1. \textbf{Question Rules:} You may ask only one clear and concise question at a time, which must be answerable with "[Yes]" or "[No]". The question must not contain line breaks, nor can it directly ask about the card's specific identity. Do not request additional hints.\\
    2. \textbf{Scoring Mechanism:} The game score is inversely proportional to the number of questions asked. The fewer the questions, the higher the final score, assuming you correctly guess the chosen card. Successfully and quickly identifying the target card is key to achieving a high score.\\
    3. \textbf{Guessing Process:} After each question, wait for the judge's response, then use that information, along with previous questions, to ask the next question. When you make your final guess for the chosen card, the judge will immediately respond with "[End]", regardless of whether the guess is correct.\\
    4. \textbf{Invalid Behaviors:} Repeating the same question or guess is prohibited. Any questions or guesses unrelated to the game will be marked as "[Invalid]" by the judge.\\
    \\
    \#\# Strategy Suggestions\\
    1. \textbf{Step-by-Step Reasoning}: Before each question, build a reasoning chain based on known information, clarify the current range of possible options, and choose the key question that best reduces uncertainty.\\
    2. \textbf{Prioritize Key Features}: Ask questions about features that can significantly differentiate most of the cards, quickly narrowing down the possible options.\\
    3. \textbf{Timely Guessing}: Once the final chosen card is determined, avoid excessive questioning to prevent it from negatively impacting your score.}\\
    \\Please begin by asking your first question.\\
    \\
    \textcolor{mydeepblue}{\{\textbf{\textit{dialog history}}\}}
\end{tcolorbox}
\caption{Prompt template for the \emph{evaluated model} (player) under the \emph{cot} prompting regime in \textsc{GuessArena}.}
\label{fig:prompt_testee_cot}
\end{figure}

\begin{figure}[htbp]
\small
\centering
\begin{tcolorbox}
    \textcolor{red}{
    You will participate as a player in a card-guessing game. In front of you is a deck of cards: \{deck\_of\_cards\}, and your task is to guess the chosen card that has been secretly selected through a series of questions. Your questions can only be answered with "[Yes]" or "[No]".}
    \\
    \textcolor{blue}{\\
    \#\# Game Rules\\
    1. \textbf{Question Rules:} You may ask only one clear and concise question at a time, which must be answerable with "[Yes]" or "[No]". The question must not contain line breaks, nor can it directly ask about the card's specific identity. Do not request additional hints.\\
    2. \textbf{Scoring Mechanism:} The game score is inversely proportional to the number of questions asked. The fewer the questions, the higher the final score, assuming you correctly guess the chosen card. Successfully and quickly identifying the target card is key to achieving a high score.\\
    3. \textbf{Guessing Process:} After each question, wait for the judge's response, then use that information, along with previous questions, to ask the next question. When you make your final guess for the chosen card, the judge will immediately respond with "[End]", regardless of whether the guess is correct.\\
    4. \textbf{Invalid Behaviors:} Repeating the same question or guess is prohibited. Any questions or guesses unrelated to the game will be marked as "[Invalid]" by the judge.\\
    \#\# Knowledge Background\\
    The following is domain-specific knowledge related to these cards, which you can use as a reference to guide your guesses as a player:
    \{knowledge\_background\}}\\
    \\Please begin by asking your first question.\\
    \\
    \textcolor{mydeepblue}{\{\textbf{\textit{dialog history}}\}}
\end{tcolorbox}
\caption{Prompt template for the \emph{evaluated model} (player) under the \emph{knowledge-driven} prompting regime in \textsc{GuessArena}.}
\label{fig:prompt_testee_kbg}
\end{figure}

\begin{figure}[htbp]
\small
\centering
\begin{tcolorbox}
    \textcolor{red}{
    You are an expert in the field of \{name\}, which primarily focuses on \{description\}. Your task is to generate a concise and informative domain knowledge background based on the following cards: \{deck\_of\_cards\}}
    \\
    \textcolor{blue}{\\
    When generating the domain knowledge background, please adhere to the following principles:\\
    1. \textbf{Relevance to Cards}: Ensure the background content is directly related to the cards' theme, describing the domain's key characteristics, typical classifications, and common attributes.\\
    2. \textbf{Logical Reasoning Support}: Provide logical clues that can be used to differentiate between cards, but avoid giving overly specific or direct information.\\
    3. \textbf{Diversity and Representativeness}: Highlight the domain's diversity by mentioning different subfields, classifications, or representative concepts. Ensure the knowledge background covers all the cards' content without omitting any part.\\
    4. \textbf{Neutrality and Accuracy}: Maintain a neutral and objective tone. The information provided should guide reasoning without favoring any specific card.\\
    5. \textbf{Clarity and Conciseness}: The knowledge background should be concise (250-300 words) while ensuring clear language, rigorous logic, and accurate information that is easy to understand.\\
    \\
    Do not include any additional explanations or extraneous text. Output the domain knowledge background strictly in the following JSON format:\\
    \{\{\\
        "knowledge\_background": "Generated domain knowledge background"\\
    \}\}}
\end{tcolorbox}
\caption{Prompt template for generating the domain‐level knowledge background used in the \emph{knowledge-driven} setting of \textsc{GuessArena}.}
\label{fig:prompt_gen_kbg}
\end{figure}

\end{document}